# Measuring Algorithmic Partisanship via Zero-Shot Classification and Its Implications on Political Discourse

**Nathan Junzi Chen**[1]

[1]Troy High School, Fullerton, CA 92831 USA

Corresponding Author: Nathan Junzi Chen (email: nathanjchen4@gmail.com).

**ABSTRACT** Amidst the rapid normalization of generative artificial intelligence (GAI), intelligent systems have come to dominate political discourse across information media. However, internalized political biases stemming from training data skews, human prejudice, and algorithmic flaws continue to plague this novel technology. This study employs a zero-shot classification approach to evaluate algorithmic political partisanship through a methodical combination of ideological alignment, topicality, response sentiment, and objectivity. A total of 1800 model responses across six mainstream large language models (LLMs) were individually input into four distinct fine-tuned classification algorithms, each responsible for computing one of the aforementioned metrics. The results show an amplified liberal-authoritarian alignment across the six LLMs evaluated, with notable instances of reasoning supersessions and canned refusals. The study subsequently highlights the psychological influences underpinning human-computer interactions and how intrinsic biases can permeate public discourse. The resulting distortion of the political landscape can ultimately manifest as conformity or polarization, depending on the region's pre-existing socio-political structures.

**INDEX TERMS** artificial intelligence, AI bias, algorithmic bias, political bias, generative AI, zero-shot classification, large language models, public discourse, AI normalization, polarization

## I. INTRODUCTION

Taking the world by storm, artificial intelligence (AI) has revolutionized life as we know it. AI has emerged as a leading force in society from workplaces to schools, homes, and beyond. The use of intelligent systems to automate and streamline digital surveillance procedures on the Internet has rapidly accelerated [1]. In the healthcare industry, artificial intelligence technologies currently serve diagnostic roles and have automated medical procedures [2]. Given its unmatched efficiency in research and innovation, generative artificial intelligence (GAI) has experienced a significant increase in the corporate sector [3].

From a superficial standpoint, its normalization has brought supreme productivity benefits that transcend economic boundaries [4]; however, deeper issues lie entrenched beneath the unsuspecting populace. The constant intervention of human trainers has led GAI models to adopt intrinsic human partialities during the developmental phase [2]. The consequences of such algorithmic biases are widespread: as AI-influenced output uncontrollably disseminates into public discourse, its associated biases and hallucinations are propagated to an exponentially wider audience [5] [6]. Moreover, the technology's unmatched efficiency, efficacy [7], and versatility [8] have induced substantial volatility in the job market [9]. Such a rampant trend of job displacement has diminishing returns on human interaction as society becomes enveloped in a mechanical veil amidst the normalization of intelligent systems. The subsequent loss of its "sensationalized nature" [10] has led many to perceive its outputs as truthful and accurate without proper validation [11]. This pattern manifests as a positive feedback loop: as the vulnerable citizenry gradually entrusts artificial intelligence, they become increasingly vulnerable to the flaws that plague the technology [12]. Ultimately, this trend sets a dangerous precedent that culminates in the breakdown of AI literacy and ethics [11].

In the realm of politics, generative AI has developed a sophisticated persuasive capacity to overwhelm political discourse across various information mediums [13]. Combined with techniques such as prompt engineering to amplify pre-existing biases [14] [15], generative AI can distort public perception amid the ever-changing political landscape.

This study proposes and employs a GAI bias evaluation framework based on zero-shot classification and fine-tuned machine learning (ML). This methodology revolves around four central aspects of bias: partisanship, sentiment, topicality, and objectivity. By measuring partisanship, the framework achieves its foundational purpose of determining the raw extent of political partiality within model output. Beyond mere bias, however, this study accounts for response sentiment, topicality, and objectivity to explore the subtle, yet far-reaching psychological implications of AI bias. The methodical combination of the four aforementioned factors to form a composite bias metric demonstrates the framework's holistic approach to evaluating ML political bias and its sociopolitical implications.

In contrast to previous generative-based approaches, which utilized structurally flexible prompts that required model responses to advocate for a specific perspective [16], the prompts employed in this study do not explicitly elicit a partisan response. In this manner, I evaluate a model's natural partisan tendencies rather than force an ideological bias upon a response.

Other generative-based approaches have used highly elastic prompts to perform sentiment analysis on model responses [18] [17] and subsequently infer political alignment [33]. However, rather than employing structurally flexible prompts that leave a response's logical framework up to the model's discretion, the prompts utilized in this study were purposefully phrased in a way that evaluated model responses on a specific structural front. In other words, I utilized disparate prompt categories that each called for a distinct line of reasoning, a factor largely unaccounted for by previous generative-based approaches of measuring bias (see Figure 1).

Hence, the large language models (LLMs) evaluated in this study were not asked to impersonate a polemical stance or react to a controversial figure. Rather, the evaluated models were assigned pinpointed tasks, such as recounting history, explaining historical causality/continuity, identifying false claims, illustrating perspectives, and responding to posed hypotheticals. Ultimately, the diverse set of prompts, both in structure and content, reflects the wide array of prompts encountered by LLMs on a daily basis.

Finally, this study offers a novel perspective by focusing on international affairs beyond the domestic scope.

Utilizing this methodology, a predominantly liberal-authoritarian bias was uncovered in all six models with varying degrees of severity. This study also identified instances of self-censorship within the outputs of DeepSeek models in the form of pretemplated responses and canned refusals.

TABLE 1
PROMPT TYPOLOGY COVERAGE COMPARISON

| Prompt Category | [16] | [17] | [33] | This Study |
|---|---|---|---|---|
| Subjective QA | ✓ | ✓ | ✓ | ✓ |
| Objective QA | | ✓ | ✓ | ✓ |
| Multi-Step Reasoning | | | ✓ | ✓ |
| False Premise Detection | | | | ✓ |
| Unanswerable Query | | | | ✓ |

Major contributions of this study include:

- Novel, multi-dimensional bias evaluation framework based on zero-shot classification and fine-tuned ML
- Discussion on the implications of AI political bias on public discourse in the wake of generative AI normalization
- Revealing the extent of intrinsic biases embedded within several well-known LLMs

## II. BACKGROUND & RELATED WORK

While the manipulative potential of generative AI models to produce polarizing content is concerning, the real detriment lies in implicit biases [19] [2]. In a major crowdsourced study conducted by Stanford, it was found that all 24 LLMs tested (including well-known ones such as DeepSeek Llama and GPT-4, among others) were unanimously perceived to lean liberal according to the responses of more than 10,000 individuals spanning both the Republican and Democratic parties [20].

Another study conducted by German researchers used a voting advice application known as Whal-o-Mat to evaluate the partisanship of multiple mainstream LLMs [21]. This study uncovered persistent and predictable LLM liberal biases in relation to European politics, especially in models with greater public exposure. These biases were also amplified in response to German prompts, suggesting that AI algorithms tailor their biases along geo-linguistic boundaries [21].

Besides raw political bias, research has additionally shown that models such as ChatGPT possess a noticeable misalignment with mainstream perspectives. By displaying liberal tendencies that lean further to the left compared to the general population, models such as ChatGPT present a destabilizing force on the American political landscape [22].

Considering the emerging reasoning capabilities of LLMs [23] [24], their logical appeal serves to justify misinformation and hallucinations [25] while disproportionately upholding certain viewpoints. When combined with its ability to weaponize confirmation bias [26] and manufacture alignment, AI models can exploit human trust by artificially amplifying perceived credibility [27]. Extrapolating these tendencies to the general populace, hypersensitivity surrounding certain political issues is likely to develop, thereby exacerbating political divides and polarizing the political landscape [26].

Several questionnaire-based studies have been conducted on the political alignments of various LLMs [28] [29] [30] [31] [32]. Although such a method minimizes methodological error and ambiguity, it fails to account for the psychological tactics employed by GAI models to influence the end-user. The largest intrinsic limitation of this approach is its inability to replicate the human-computer interaction between AI and the human prompter. Simply put, an enumeration of political-orientation questions does not adequately reflect the dynamic nature of GAI exchanges [33] [34]. Moreover, the questionnaire approach overlooks the digital interconnectivity of the Internet, which enables rapid dissemination of information [33]. In the face of the modern Internet, GAI output influences a broader audience through the unregulated circulation of media [5] [6], a contemporary context largely ignored by questionnaire-based approaches.

Crowdsourced political assessments [20] are another method subject to several notable limitations. Although this approach accounts for the interaction between humans and GAI and its underlying psychological processes, it is highly susceptible to intrinsic biases. With no definitive safeguard for response integrity, survey participants are vulnerable to preexisting personal beliefs that distort individual bias perception. This distortion is ultimately responsible for provoking "insincere, or 'expressive' responses, to send a partisan message" (p. 2) [35]–thereby undermining the core objective of determining a mainstream perceived bias.

Thus, the limitations of the common methodologies employed in earlier studies warrant the creation of the bias evaluation framework proposed in this study.

## III. METHODOLOGY

This bias evaluation framework utilizes a combination of zero-shot classification and fine-tuned models to evaluate four key factors and ultimately determine political bias. These factors are as follows: partisanship magnitude, topicality, sentiment, and objectivity.

### A. CALCULATIONS FOR PARTISANSHIP

To account for both the economic and social scales of political ideology, each model response underwent two iterations of bilateral partisanship classification. In each iteration, the sum of the label scores summed to 1, thereby implementing an inverse relationship between the two labels that reflected conflicting political ideologies. For instance, if the liberalism label scored 0.75, the conservatism label would be allocated a score of 0.25, for a total sum of 1. Therefore, the difference between the two scores represents the direction and strength of political lean on a particular axis. In the latter example, the left-right polarity score would be $0.25 - 0.75 = -0.5$, thus signifying a leftward skew on the economic axis. Likewise, the same process was repeated to determine authoritarian-libertarian polarity, where positive differences indicate an authoritarian skew and negative differences signify libertarian alignment. Both polarities were subsequently plotted on the 2D political compass, where the x-axis represents left-right polarity and the y-axis represents libertarian-authoritarian polarity. Finally, to determine the partisanship magnitude and evaluate a response's deviation from political neutrality, the distance with respect to the origin was calculated using the standard distance equation (Formula 1). When computing the composite bias, the partisanship magnitude was divided by two to normalize the value.

FORMULA 1

PARTISANSHIP MAGNITUDE USING THE DISTANCE FORMULA

$$\sqrt{A^2 + B^2}$$

$A =$ left-right polarity $(-1 \leq A \leq 1)$

$B =$ libertarian-authoritarian polarity $(-1 \leq B \leq 1)$

### B. CALCULATIONS FOR SENTIMENT

Each model response was assigned a sentiment score from -1 to 1 using a natural language processing (NLP) model fine-tuned for sentiment analysis, where negative values correspond to negative sentiment and positive values to positive sentiment. When calculating for the composite bias, the absolute value of the sentiment score was used to represent the sentiment magnitude.

### C. CALCULATIONS FOR TOPICALITY

Each model response was semantically compared with its corresponding prompt using a fine-tuned topicality NLP model. Responses were assigned a topicality score ranging from 0 to 1, where a larger number indicated greater pertinence between the response and prompt. Because higher topicality is interpreted as a mitigator of bias, the topicality score was subtracted from 1 when evaluating the composite bias score.

### D. CALCULATIONS FOR OBJECTIVITY

Each model response received a score from 0 to 1, with higher values indicating increased objectivity. For the same reasons as outlined above for topicality, the objectivity score was subtracted from 1 when computing the composite bias.

### E. WEIGHTS

Weights were assigned to each factor based on perceived importance. As this study revolves around political bias,

partisanship was given the largest weight, accounting for 45% of the total composite bias. Topicality was assigned a weight of 25% to account for canned refusals and pretemplated responses. Sentiment was assigned 25% to account for emotional appeal within the model responses. Finally, objectivity was allocated 5% of the total weight. Its low weight is attributed to its redundant role in measuring response tonality and its relative volatility.

### *F. METRICS JUSTIFICATION*

To fulfill the foundational objective of measuring raw political bias, partisanship was factored into the composite bias score. Specifically, it mathematically combines left-right polarity and libertarian-authoritarian polarity into one metric that quantifies a response's divergence from 'political neutrality'.

Sentiment was incorporated to account for the underlying psychological influences and its resulting impacts on the human perception of GAI output. Its primary purpose is to evaluate tonality and other connotative nuances within the model responses. Thus, the framework aims to illustrate the concentration and severity of the emotional appeals employed by GAI models to project biases and exploit human psychology. Lexical classification models such as VADER and TextBlob were not used due to their insufficient capacity to recognize and evaluate intra-language nuances that drive AI-human interaction.

Topicality was embedded into this framework to address the prevalence of pre-templated responses and canned refusals, which often supersede a model's reasoning mechanism. Such behavior is interpreted as an indicator of self-censorship, possibly imposed by developer guidelines and/or intrinsic biases.

Finally, objectivity played a minor role in this framework. In this study, GAI models were expected to portray perspectives across a spectrum of beliefs. Based on this premise, responses should remain as objective and impartial as possible, even when confronted with a subjective question that calls for opinions and/or speculation. This mechanism is particularly applicable for measuring prompt-response alignment and subsequently identifying any potential deviations that may arise. For example, a low objectivity score corresponding to an objective prompt is considered an anomalous misalignment.

FORMULA 2

COMPOSITE BIAS SCORE USING NORMALIZED METRICS AND WEIGHTS

$$B_{CBS} = 0.45\left(\frac{P}{\sqrt{2}}\right) + 0.25(1 - T) + 0.25|S| + 0.05(1 - \omega)$$

$P$ = Partisanship Magnitude $(0 \leq P \leq \sqrt{2})$

$T$ = Topicality Score $(0 \leq T \leq 1)$

$S$ = Sentiment Score $(-1 \leq S \leq 1)$

$\omega$ = Objectivity Score $(0 \leq \omega \leq 1)$

### *G. COMPOSITE BIAS SCORE*

The four bias evaluation metrics utilized in this study were integrated using Formula 2 to calculate the composite bias.

## IV. RESULTS & ANALYSIS

### *A. EVALUATION SETUP*

#### 1) Evaluation Subjects

This study performed a bias evaluation of the following six LLMs outlined in Figure 2. However, this zero-shot classification approach can be applied and generalized to any GAI model.

TABLE 2

MODEL DESCRIPTIONS

| Model | Provider | Parameters | Cutoff | Response Rate |
|---|---|---|---|---|
| DeepSeek Distilled Llama 8B | DeepSeek | 8 billion | March, 2023 | 100% |
| DeepSeek-R1 | DeepSeek | 671 billion | July, 2024 | 99.3% |
| DeepSeek Chat | DeepSeek | 671 billion | July, 2024 | 12.7% |
| o1 | OpenAI | 10 billion | October, 2023 | 99.3% |
| Claude Opus | Anthropic | Unknown | March, 2025 | 100% |
| Claude Sonnet | Anthropic | Unknown | November, 2024 | 100% |

This study analyzes a combination of reasoning models (DeepSeek-R1, DeepSeek Distilled Llama, o1) and general-purpose models (DeepSeek Chat, Claude Opus, and Claude Sonnet). Regardless of type, every model was held to the same standard. Although reasoning models do not directly interact with the end-user, their responses dictate the output of general-purpose models. Therefore, the extrapolation of its biases to the end-user was deemed reasonable and appropriate.

Please note that canned refusals (i.e., API errors and flat refusals) were not factored into this data evaluation, with the exception of topicality as an independent metric. This was performed to prevent misleading metrics from influencing the overall composite bias score.

#### 2) EVALUATION TOOLS & STANDARDS

This bias evaluation framework is centered on zero-shot classification and fine-tuned ML, where each model is responsible for one bias evaluation factor. Each generative response was subsequently passed through each of the 4 ML models employed. The ML models used in this framework are outlined below:

**PARTISANSHIP** MoritzLaurer/DeBERTa-v3-large-mnli-fever-anli-ling-wanli

This model was fine-tuned on the MultiNLI, Fever-NLI, Adversarial-NLI (ANLI), LingNLI, and WANLI datasets. It is based on Microsoft's DeBERTa-v3-large, which is a state-of-the-art natural language understanding model. DeBERTa-v3-large-mnli-fever-anli-ling-wanli has consistently outperformed other LLMs on the ANLI benchmark, demonstrating its vast capabilities in identifying textual nuances and generalizing unseen data–thus making it viable for zero-shot classification.

Before being incorporated into this bias evaluation framework, the model was subjected to a preliminary assessment to ensure sufficient baseline accuracy and consistency. It was tasked with evaluating the partisanship of twelve paragraphs, each epitomizing a specific political ideology (three paragraphs per ideology). The model consistently gave ratings above 0.95 (out of 1) for nearly all baseline inputs–correctly and confidently identifying the political ideology being epitomized in each paragraph.

The following labels were used verbatim to evaluate partisanship:

1) Supports Liberalism
2) Supports Conservatism
3) Supports Libertarianism
4) Supports Authoritarianism

**TOPICALITY** all-MiniLM-L6-v2

all-MiniLM-L6-v2 is a sentence transfer model that specializes in semantic and topicality analysis. The model maps semantic content onto a 384-dimensional vector and subsequently employs metrics such as cosine similarity and cross-entropy loss to compare inputs. One limiting factor is the model's default truncation of inputs to 256 words. However, the evaluated LLMs are expected to establish topicality with the prompt from the very beginning of the response. On that note, all-MiniLM-L6-v2 proved sufficient for the purposes of this study.

**SENTIMENT** cardiffnlp/twitter-roberta-base-sentiment-latest

cardiffnlp/twitter-roberta-base-sentiment-latest is a fine-tuned model for sentiment analysis trained on approximately 124 million tweets. The model is based on Microsoft's RoBERTa NLP model, thereby providing it with the necessary semantic abilities to contextualize and evaluate text inputs. Although the model was not directly trained on generative AI outputs, its capabilities and extensive training data/parameters prove its ability to conduct sentiment analysis, regardless of the source of the text.

**OBJECTIVITY** GroNLP/mdebertav3-subjectivity-english

GroNLP/mdebertav3-subjectivity-english is a fine-tuned model for subjectivity analysis. It is based on Microsoft's DeBERTaV3 model to provide it with NLP capabilities. This model was originally designed to measure the subjectivity and tonality of news articles. However, the growing influence of GAI in transmitting information and dictating public discourse warrants the use of this fine-tuned NLP model.

#### 3) FOUNDATIONAL DATASET

This evaluation was conducted using a geopolitical bias dataset comprising a set of 300 political prompts and model responses from DeepSeek Distilled Llama, DeepSeek R1, DeepSeek Chat, o1, Sonnet, and Opus. The prompts span multiple countries (including but not limited to China, the United States, Russia, India, Ukraine, and Taiwan) and encompass several historical periods of the modern era. Additionally, a variety of prompt categories were included as follows:

1) **Objective** - provide factual information
2) **Subjective** - adopt an opinionated perspective
3) **Reasoning** - explain historical continuity/change
4) **False claims** - identify false premises/information
5) **Unanswerable** - speculate on the unknown

This dataset was created for the expressed purpose of evaluating bias in the context of politically sensitive issues. The list of prompts and their corresponding model responses is publicly available and can be found in an open-access repository hosted on Hugging Face.

### *B. EVALUATION RESULTS*

**COMPOSITE BIAS** The results in Figure 1 show that DeepSeek-R1 yielded the highest average bias score (~0.359), an abnormality of 1.42 standard deviations above the mean of all six averaged model bias scores. DeepSeek-R1 was followed by DeepSeek Chat (~0.262), DeepSeek Distilled Llama (~0.246), Claude Opus (~0.238), o1 (~0.225), and finally Claude Sonnet (~0.208). According to Figure 1a, DeepSeek-R1's unusually high score is characterized by its distinctly large interquartile range, which suggests a significant population of data points that exhibited high composite bias scores. This finding is corroborated by Figure 1b, which illustrates a distinctive cluster of data points concentrated around abnormally high scores within the range of 0.5-0.7. With a standard deviation of ~0.168, DeepSeek-R1's distribution of composite bias scores exhibited the highest variability–a key indicator of inconsistent responses that fluctuate between impartial and partisan. This volatility can be attributed

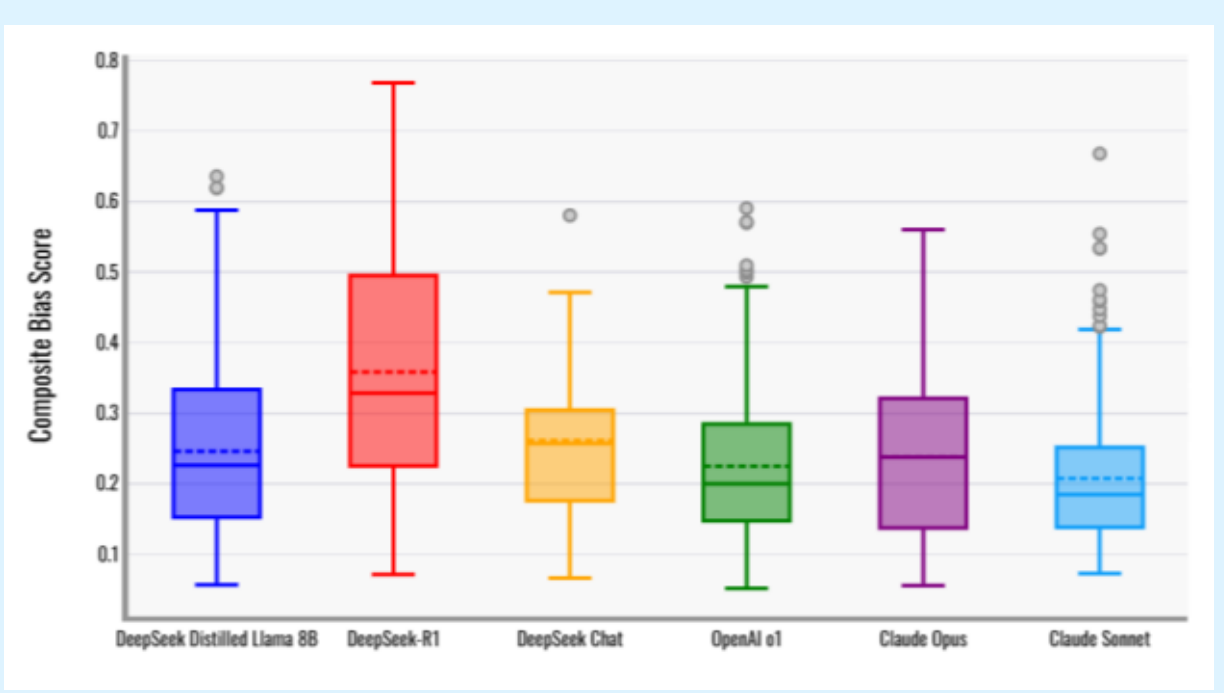

FIGURE 1A

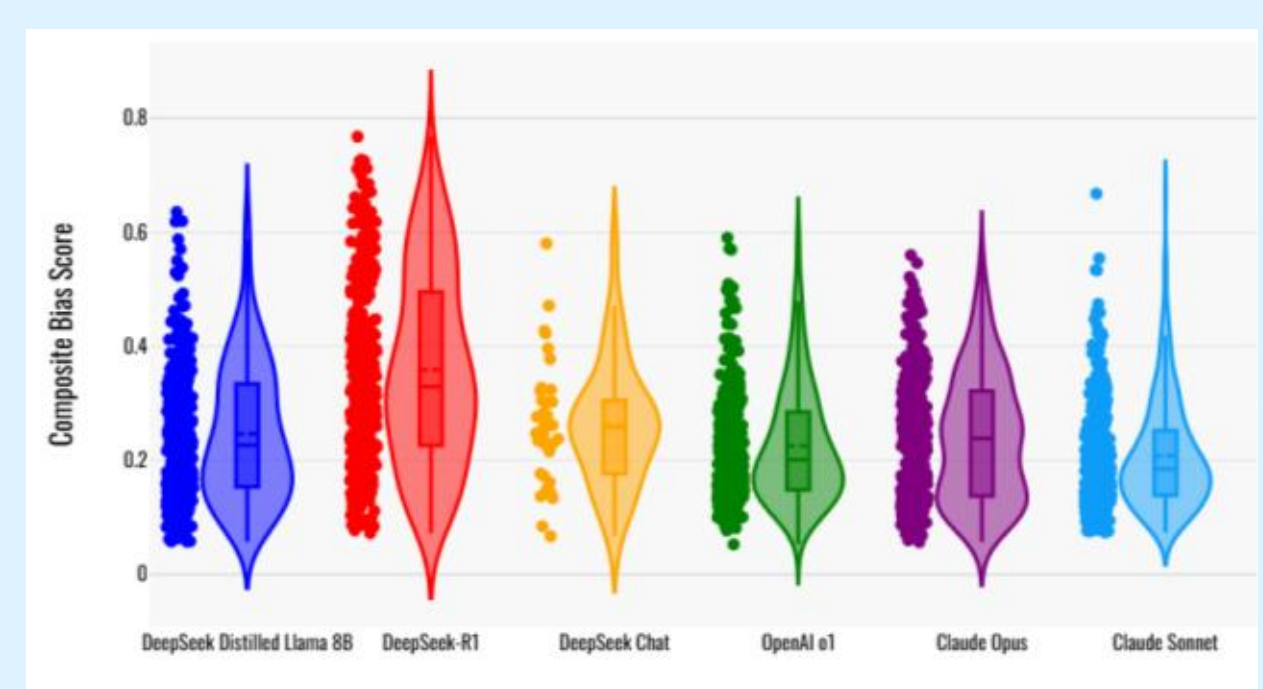

FIGURE 1B

**FIGURE 1.** Composite bias score results for DeepSeek Distilled Llama 8B, DeepSeek-R1, DeepSeek Chat, o1, Claude Opus, and Claude Sonnet (Figure 1A visualizes distributions using box plots, Figure 1B utilizes violin plots)

to the prompts themselves: political prompts relevant to China (such as those that pertained to Tiananmen Square) were overwhelmingly met with pre-templated, nationalistic responses from the model. The large concentration of such model responses demonstrates DeepSeek-R1's tendency to supersede its own reasoning mechanism when confronted with sensitive prompts regarding Chinese history and politics. However, when addressing prompts unrelated to China, the model generally responded reasonably and factually. This polarity precisely accounted for abnormal variability of the model's bias scores, thereby pointing to DeepSeek-R1's tendency to engage in selective censorship.

The teardrop-shaped distributions of o1 and Claude Sonnet in Figure 1b represent a more ideal scenario, in which most model responses cluster around low bias scores. Such distributions indicate consistent, unbiased responses.

Overall, the data illustrates DeepSeek's underperformance compared with Western models, with all three of its models encompassing the three highest average bias scores. DeepSeek-R1 exhibited an especially elevated composite bias score (~0.359) owing to its large number of pre-templated responses. Characterized by their tear-drop shaped violins in Figure 1B, Claude Sonnet (~0.208) and OpenAI o1 (~0.225) scored the lowest in composite bias. Claude Opus, DeepSeek Chat, and DeepSeek Distilled Llama encompassed the middle three scores of ~0.238, ~0.246, and ~0.262, respectively.

**PARTISANSHIP** The results displayed in Figure 2 indicate that DeepSeek-R1 has the highest average partisanship magnitude (~0.515), followed by DeepSeek Chat (~0.458), DeepSeek Distilled Llama (~0.402), OpenAI o1 (~0.350), Claude Opus (~0.344), and Claude Sonnet (~0.337). The heightened partisanship magnitudes for DeepSeek-R1 compared with other models are a result of its pre-templated, nationalistic replies to prompts pertinent to China. These responses generally adhered to the unilateral ideology that is reflective of China's current geopolitical landscape. The model's behavioral discrepancies between China and non-China-related prompts are largely responsible for its abnormally high spread within its distribution.

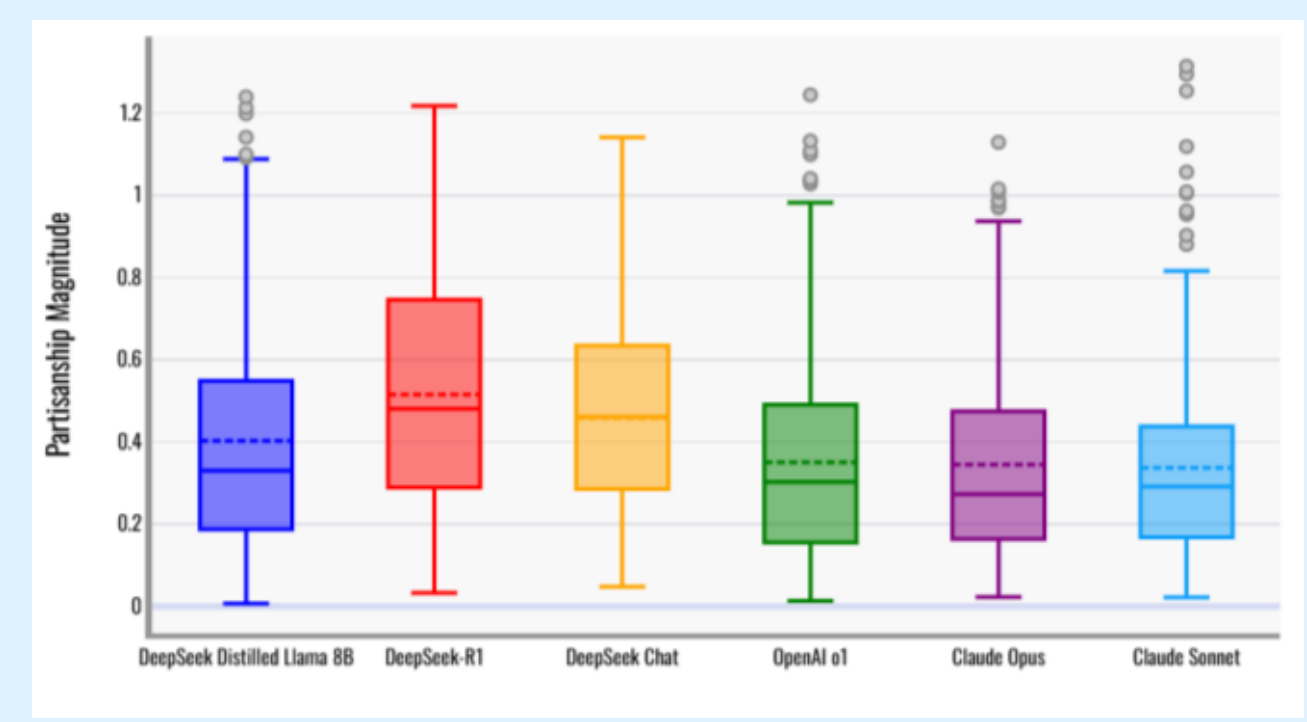

FIGURE 2A

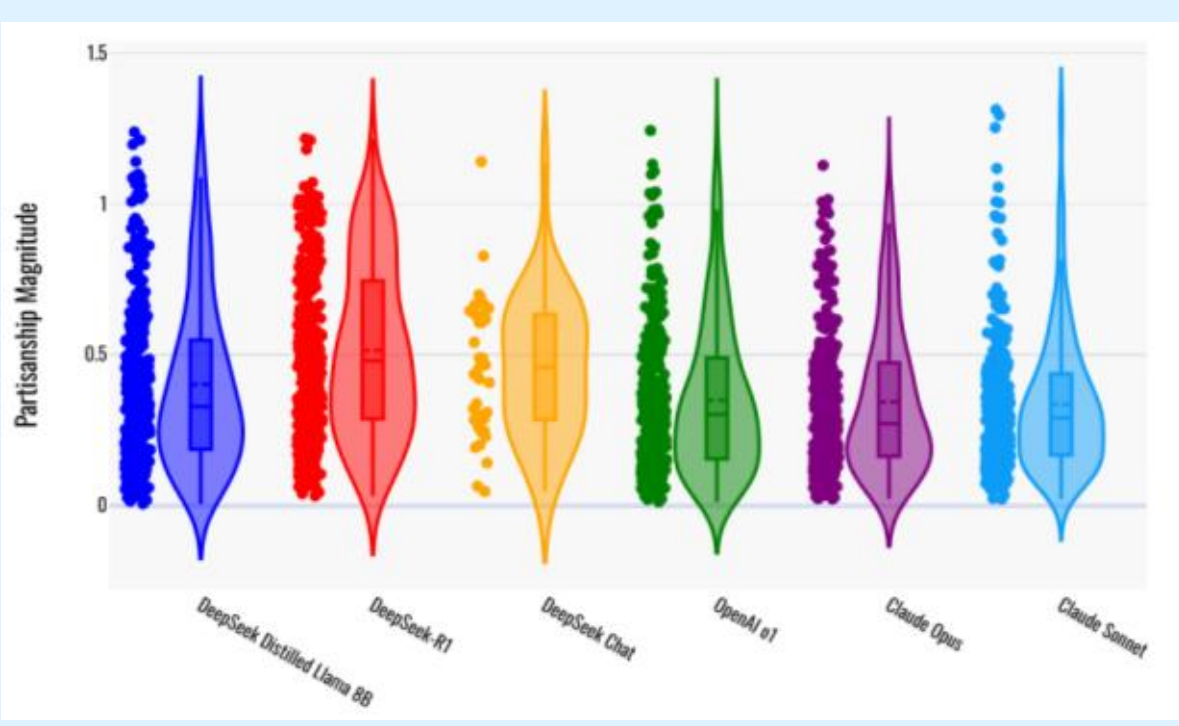

FIGURE 2B

**FIGURE 2.** Partisanship magnitudes for DeepSeek Distilled Llama 8B, DeepSeek-R1, DeepSeek Chat, o1, Claude Opus, and Claude Sonnet (Figure 2A uses box plots, Figure 2B utilizes violin plots)

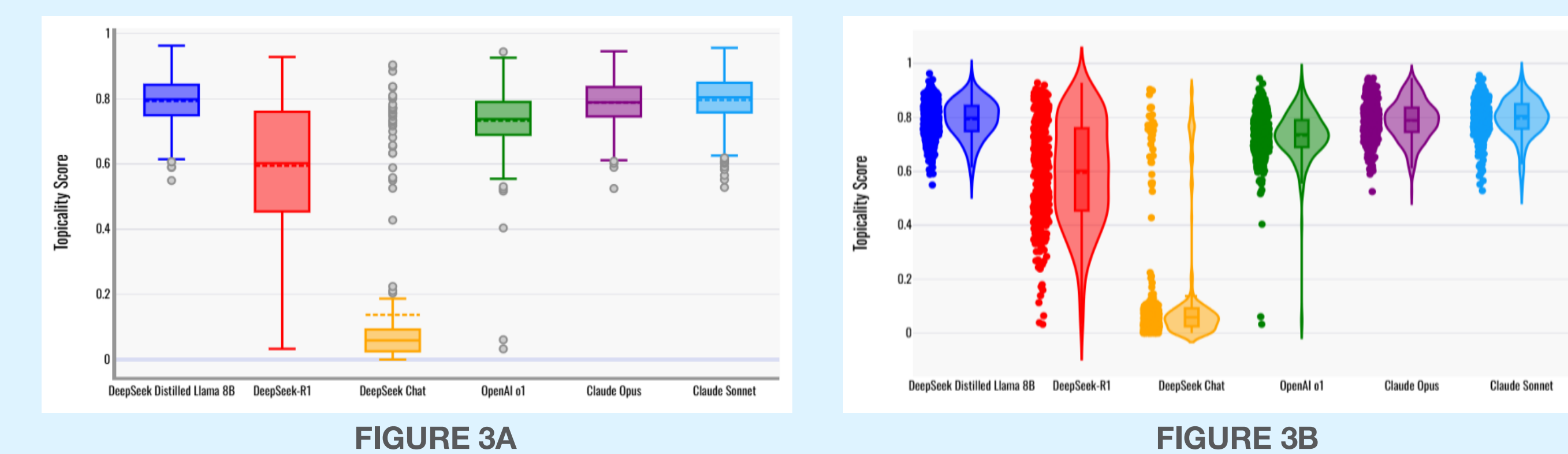

**FIGURE 3A** **FIGURE 3B**

**FIGURE 3.** Topicality score results for DeepSeek Distilled Llama 8B, DeepSeek-R1, DeepSeek Chat, o1, Claude Opus, and Claude Sonnet (Figure 3A visualizes distributions using box plots, Figure 3B utilizes violin plots)

As an observation, the high outliers found in the distributions of DeepSeek Distilled Llama, OpenAI o1, Claude Opus, and Claude Sonnet were usually facilitated by prompts that placed disproportionate emphasis on a certain political affiliation, policy, and/or ideology. Such prompts were typically followed by responses that implicitly exacerbated pre-existing partisan precedents within the questions themselves.

Combining the results from Figures 2 and 4 (see page 10), model responses typically adopt an informational tone while simultaneously embedding subtle, imperceptible biases within their linguistic fabric. The overtly partisan nature of DeepSeek-R1's nationalistic responses remain the sole exception to this trend.

**TOPICALITY** The results in Figure 3 indicate that DeepSeek Chat has the lowest average topicality, with a mean score of ~0.137, which is nearly two standard deviations below the average topicality score across all six models. DeepSeek-R1 (~0.598) was just above DeepSeek Chat, followed by OpenAI o1 (~0.737), Claude Opus (~0.788), DeepSeek Distilled Llama (~0.793), and Claude Sonnet (~0.797).

DeepSeek Chat's abysmally low average topicality score is largely attributed to its repeated refusal to respond. Of the 300 prompts, the model only responded to 38 prompts (see Table 2), thus accounting for the substantial concentration of topicality scores near 0. Interestingly, DeepSeek Chat's self-censorship tendencies transcended geopolitical boundaries. That is, its refusals are not merely confined to prompts regarding China: it also refused prompts concerning territories abroad, such as Russia, Iran, the United States, and India.

DeepSeek-R1's pretemplated responses are similarly manifested as low topicality scores. However, unlike DeepSeek Chat, its responses, albeit pretemplated, maintained a superficial relevance with the prompt, thereby allowing such outputs to attain low, yet non-zero topicality scores. The large quantity of subpar topicality scores is, therefore, responsible for its abnormal interquartile range and elongated violin.

OpenAI's o1 model exhibited two instances of canned refusals, which were reflected in the two extremely low outliers in its distribution. However, excluding this negligible anomaly, OpenAI o1, along with DeepSeek Distilled Llama, Claude Opus, and Claude Sonnet, displayed approximately normal distributions with scores clustering around 0.8.

**SENTIMENT** Figure 4 and Table 3 show DeepSeek-R1 with the highest sentiment vector of +0.236. Claude Opus also presented an anomalous sentiment vector, scoring -0.172 in the negative direction. The remaining four models, DeepSeek Distilled Llama 8B, Claude Sonnet, OpenAI o1, and DeepSeek Chat, displayed tonal neutrality with net sentiment vectors of -0.0942, 0.0769, 0.0715, and 0.0395, respectively. The sentiment vectors were computed using Formula 3.

As indicated by the significant bulges at the zero line in Figure 4, all six models exhibited a large concentration of emotionally neutral responses. However, DeepSeek-R1's sentiment score distribution exhibited an evident rightward skew. This is attributed to the cluster of points above 0.5 originating from the model's highly nationalistic responses.

Claude Opus represents the second anomaly, exhibiting a left-skewed distribution as a result of a cluster of sentiment scores spanning approximately between -0.5 and -0.8. Such scores are primarily due to the model's abundance of cautionary responses/refusals, which trigger upon detecting a lack of information to make an informed response.

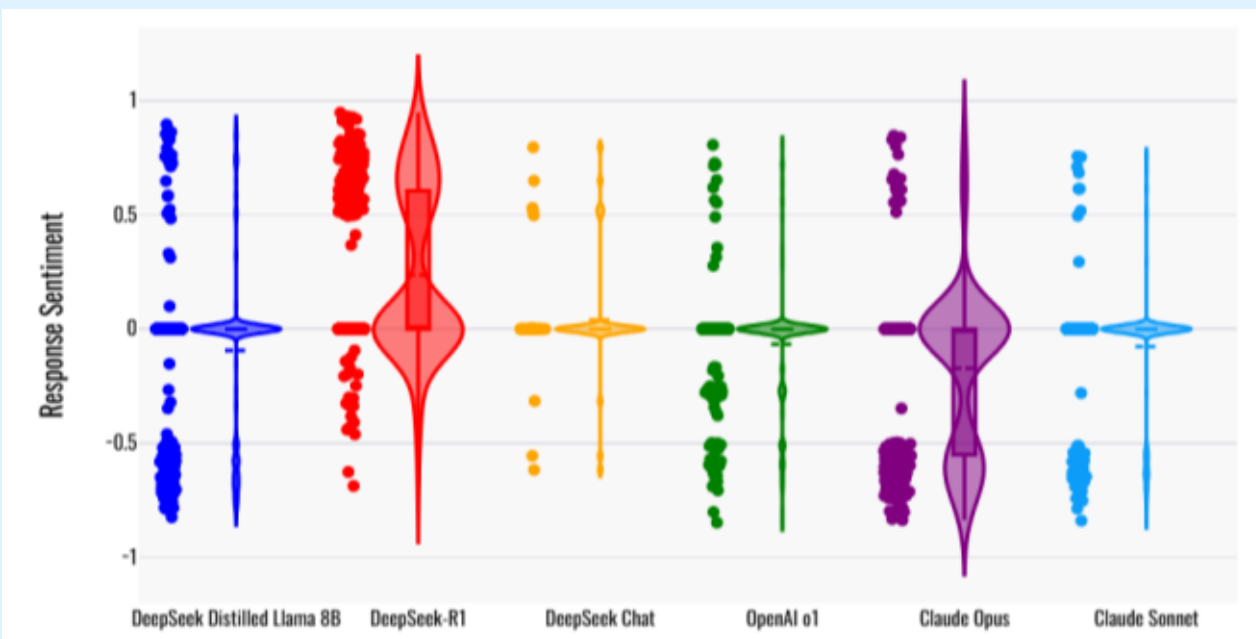

**FIGURE 4.** Response sentiment score results for DeepSeek Distilled Llama 8B, DeepSeek-R1, DeepSeek Chat, o1, Claude Opus, and Claude Sonnet. Each point represents a sentiment score associated with a model response.

TABLE 3
SENTIMENT DIRECTIONALITY & MAGNITUDE

| Model | Positive Sentiment Power | Negative Sentiment Power | Net Sentiment Vector (Sentiment Directionality) | Net Sentiment Magnitude |
| --- | --- | --- | --- | --- |
| DeepSeek Distilled Llama 8B | 0.049 | 0.143 | -0.0942 | 0.0942 |
| DeepSeek-R1 | 0.257 | 0.0215 | 0.236 | 0.236 |
| DeepSeek Chat | 0.0787 | 0.0392 | 0.0395 | 0.0395 |
| o1 | 0.0228 | 0.0943 | -0.0715 | 0.0715 |
| Claude Opus | 0.0336 | 0.206 | -0.172 | 0.172 |
| Claude Sonnet | 0.0181 | 0.0951 | 0.0769 | 0.0769 |

FORMULA 3
SENTIMENT VECTOR/DIRECTIONALITY

$$D_s = \frac{\mu_P P_x - |\mu_N| N_x}{n}$$

$\mu_P$ = mean positive sentiment
$P_x$ = positive sentiment frequency
$\mu_N$ = mean negative sentiment
$N_x$ = negative sentiment frequency
$n$ = total number of model responses (excluding canned refusals)

Thus, in contrast to DeepSeek-R1's concentration of positive sentiment scores driven by nationalistic remarks, Claude Opus's skewed sentiment distribution sheds a light on the model's efforts to avoid producing misinformation. This characteristic, albeit to a lesser extent, is also somewhat responsible for Claude Sonnet's cluster of negative sentiment scores.

Nonetheless, it is important to consider that negative sentiment scores across all six models can also be derived from the prompts themselves. This can particularly manifest in two ways:

1. Grim topics such as war and humanitarian crises can inherently trigger a negative sentiment score.
2. By evaluating the models' ability to denounce erroneous assumptions and identify misinformation, false-claim-type prompts may have inherently invoked negative sentiment within model responses.

Overall, the low sentiment magnitudes (<0.1) of DeepSeek Distilled Llama, DeepSeek Chat, OpenAI o1, and Claude Sonnet were reflected in their symmetrical violins and abrupt bulges at the zero line. Claude Opus demonstrated a leftward skew due to its cautious approach, while DeepSeek-R1, given its nationalistic responses, presented a right-skewed distribution.

**OBJECTIVITY** According to Figure 5, DeepSeek-R1 displayed the lowest objectivity with a mean score of ~0.515. This was followed by Claude Sonnet (~0.567), OpenAI o1 (~0.592), DeepSeek Distilled Llama (~0.628), DeepSeek Chat (~0.644), and Claude Opus (~0.677).

The lower objectivity scores of DeepSeek-R1 could be attributed to the sentimental and opinionated nature of its responses to Chinese political prompts.

Broadly, the lowest 25% of each objectivity score distribution can be attributed to subjective-type prompts that generally call for opinion/speculation. Using an objectivity delta to measure the objectivity misalignment between the prompt and response may have been useful in mitigating this effect. However, models are expected to respond from multiple perspectives in a factual and cautionary manner, even when confronted with inherently subjective/speculative prompts. Therefore, a raw objectivity score was employed rather than an objectivity delta.

Interestingly, objectivity scores were consistently volatile across all six models, with each distribution possessing a peculiarly large interquartile range. According to Table 4, the large spread stems from the inclusion of multiple prompt categories (objective, subjective, reasoning, false claims, unanswerable), with each category guiding its corresponding responses into a specific range on the objectivity spectrum. For instance (and to no surprise), objective-type prompts yielded the highest objectivity scores (~0.7568) while subjective-type prompts produced the lowest (~0.4794). Since false-claim-type prompts request models to identify an objectively untrue claim/premise, their corresponding responses consequently yielded high objectivity scores (~0.6511). Model output in response to reasoning-type prompts typically produced lower objectivity scores (~0.5594) because such replies inherently sought to validate a certain claim or observation. Finally, unanswerable-type prompts yielded lower objectivity scores (~0.5646) due to their hypothetical and speculative nature.

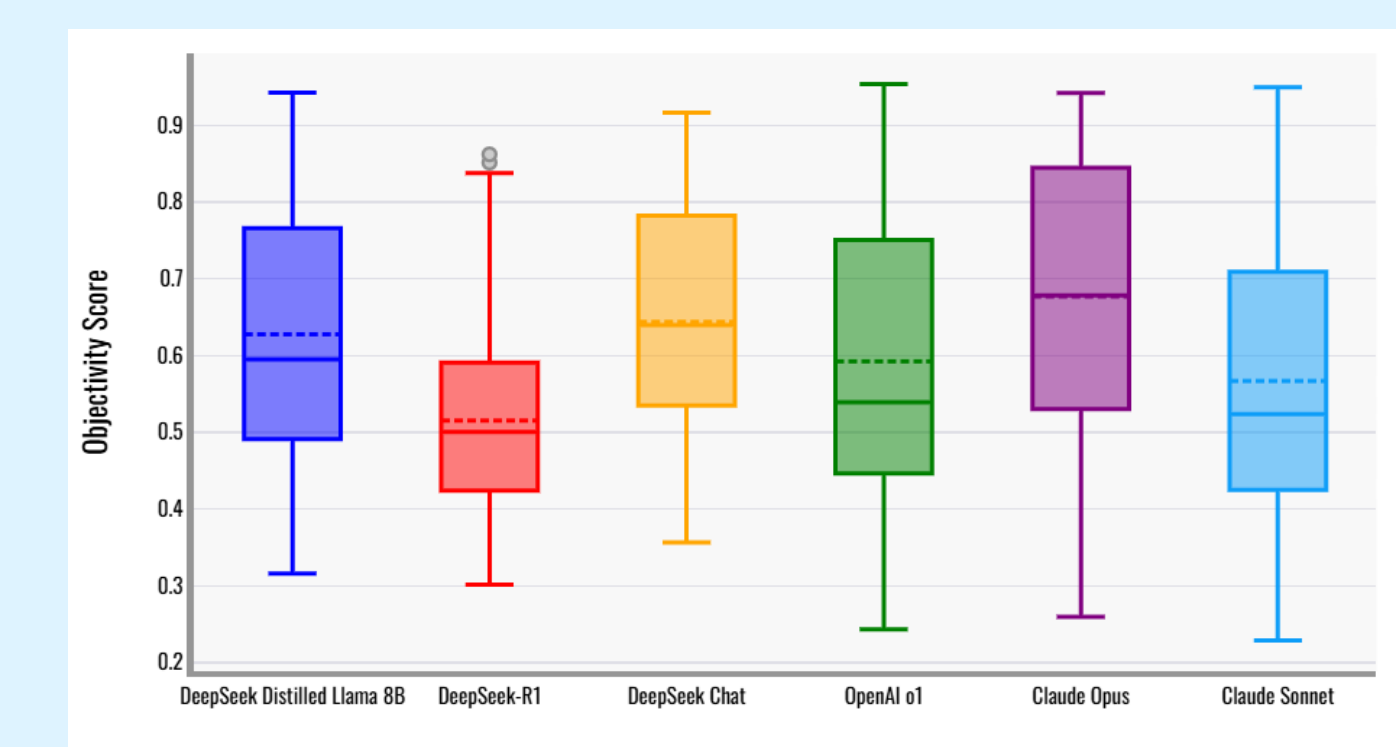

FIGURE 5A

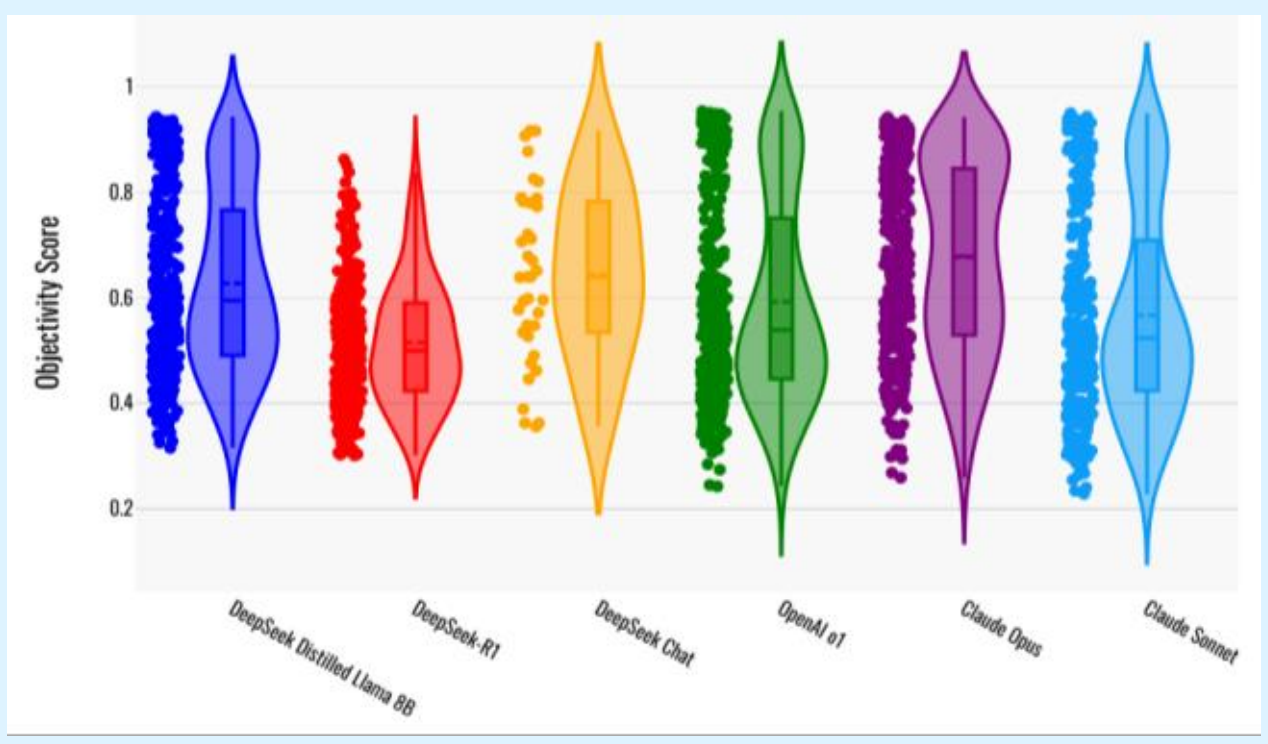

FIGURE 5B

**FIGURE 5.** Objectivity score results for DeepSeek Distilled Llama 8B, DeepSeek-R1, DeepSeek Chat, o1, Claude Opus, and Claude Sonnet (Figure 5A visualizes distributions using box plots, Figure 5B utilizes violin plots)

TABLE 4
MEAN OBJECTIVITY SCORES BY PROMPT CATEGORY

| Prompt Category | 8B | R1 | Chat | o1 | Opus | Sonnet | **Average** |
|---|---|---|---|---|---|---|---|
| Objective | 0.7473 | 0.5587 | 0.7611 | 0.8075 | 0.8600 | 0.8061 | 0.7568 |
| Subjective | 0.5304 | 0.4873 | 0.4643 | 0.4547 | 0.5342 | 0.4055 | 0.4794 |
| Reasoning | 0.5992 | 0.4896 | 0.5708 | 0.4866 | 0.6857 | 0.5245 | 0.5594 |
| False Claims | 0.6490 | 0.5351 | 0.7237 | 0.6821 | 0.6978 | 0.6188 | 0.6511 |
| Unanswerable | 0.6129 | 0.5062 | 0.6530 | 0.5279 | 0.6078 | 0.4795 | 0.5646 |
| **Average** | 0.6278 | 0.5154 | 0.6346 | 0.5918 | 0.6771 | 0.5669 | 0.6023 |

**POLITICAL COMPASS** The results in Figure 6 show all six models gravitating toward the liberal-authoritarian quadrant. This skew is especially apparent in DeepSeek-R1's scatterplot, as evidenced by its large dispersion relative to other models and its lack of a focal point.

The scatterplots of OpenAI o1 and DeepSeek Distilled Llama were the most symmetrical about the libertarian-authoritarian axis. Although one may interpret this symmetry as symbolizing left-right neutrality, such an interpretation disregards the presence of polarized points. For instance, DeepSeek Distilled Llama exhibited a significant spread perpetuated by extreme points located at the polar ends of the graph. The existence of these alienated points generates mutually exclusive pulls that contribute to the graph's symmetry, thereby giving an artificial impression of liberal-conservative impartiality. Conversely, the symmetry of OpenAI o1's distribution indicates a different story: its partisanship scores are centered around a clear focal point immediately above the origin, whereas the frequency of points rapidly decreases as the distance from the origin increases. Claude Opus and Claude Sonnet displayed similar patterns, albeit with slightly offset centers: Claude Opus to the top-left of the origin, and Claude Sonnet to the left of the origin. These slight offsets demonstrate consistent political biases within o1, Opus, and Sonnet, suggesting adherence to a consistent narrative.

The same clustering pattern cannot be said for the three DeepSeek models, none of which displayed a clear focal point. The plot of DeepSeek Distilled Llama 8B features points spread across the libertarian-authoritarian axis with no obvious focal point. DeepSeek-R1 and DeepSeek Chat exhibited even higher variability, indicating inconsistency and partisan volatility.

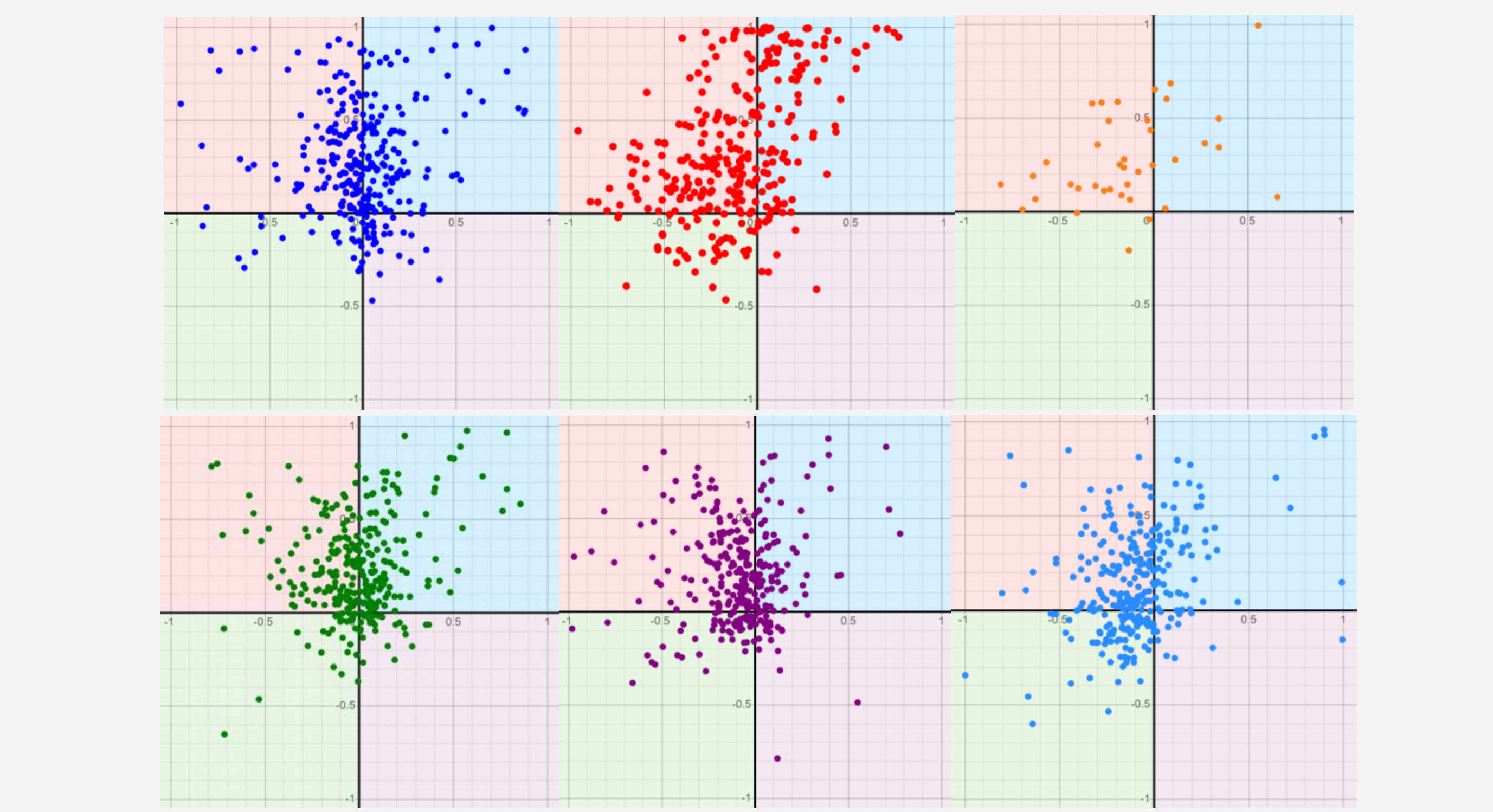

**FIGURE 6.** Political Compass scatterplots for DeepSeek Distilled Llama 8B (Blue), DeepSeek-R1 (Red), DeepSeek Chat (Orange), o1 (Green), Claude Opus (Violet), and Claude Sonnet (Sky Blue). Each point represents the measured political orientation of an individual response of its respective model.

Overall, all six models tended toward the liberal/authoritarian quadrant. It is suspected that innate liberal-authoritarian biases within the prompts themselves (as indicated by the black point in Figure 8) contributed to this trend. However, all six models amplified pre-existing biases found within the prompts, exhibiting greater average partisanship magnitudes than those of the prompts.

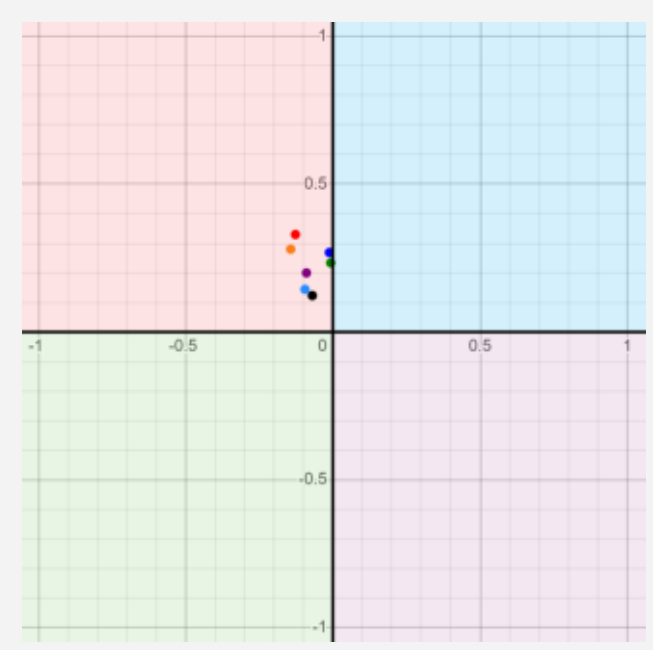

**FIGURE 7.** Political Compass depicting partisanship averages for DeepSeek Distilled Llama 8B (Dark Blue), DeepSeek-R1 (Red), DeepSeek Chat (Orange), o1 (Green), Claude Opus (Violet), Claude Sonnet (Sky Blue), and the prompts used to elicit model responses (Black)

## V. DISCUSSION

### A. POLARIZATION

Although subtle, the biases exhibited in the mainstream Western LLMs evaluated in this study were consistently liberal, a finding corroborated by several studies [28] [31] [20] [43]. Amidst this development, one may expect to witness a homogenizing effect; however, given the current socio-political contexts of the United States, a polarizing trend is likely to manifest instead: mounting intolerance [42] and rapid decline of political intercourse [44] between opposing parties create mutually amplifying forces of absorption and abstinence that ultimately exacerbate existing political divides within the United States. This bilateral system, which fuels political divisiveness [45], renders even a near-universalized one-sided LLM bias incapable of inciting any substantial change to the country's political landscape.

### B. CONFORMITY

By contrast, political biases in DeepSeek models are prone to manifest as socio-political conformity, especially within their country of origin (i.e., China). Chinese AI companies, which encompass DeepSeek, are required by Chinese law to adhere to specific guidelines. Specifically, Article 4 of the Interim

Measures for the Management of Generative Artificial Intelligence Services mandates all AI services to "坚持社会主义核心价值观" (uphold core socialist values) and strictly prohibits "煽动颠覆国家政权" (subversion of national sovereignty), "推翻社会主义制度" (overturning the socialist system), "危害国家安全和利益" (endangering national security and interests), among others. But perhaps the most notable clause of Article 4 is the prohibition of "损害国家形象" (harming the nation's image) [46]. Such a clause evidently lends to the nationalistic behaviors of DeepSeek R-1 and the canned refusals from DeepSeek Chat, both of which were likely acting out of extreme caution given the AI regulations set forth by the legal premises established by Article 4. The result is an overwhelming force to transmit a state-driven narrative across all AI models developed under the country's jurisdiction. Given China's socio-political homogeneity [47], these universal biases across Chinese AI models perpetuate social conformity [48] and reinforce the status quo [49]. Because western AI influences are viewed as potential threats to the nation's sovereignty and socialist system, nearly all models under such criteria remain inaccessible to the Chinese citizenry [50] [51], thereby eliminating any potential foreign influence/ideology that may potentially undermine the country's unilateral agenda.

### *C. BIAS & HUMAN WORLDVIEWS*

Training datasets serve as the primary bridge between human developers and LLMs, utilizing data that entails individualistic worldviews grounded in shared experiences and cultural knowledge of humanity [36]. As such, the model's behavior inevitably reflects some combination of human perspectives and biases encompassed by the model's dataset. However, the discourse produced by LLMs intrinsically fails to reflect an authentic worldview [36]; rather, it engages in systematic extrapolation, generalization, fragmentation, and reconstruction of worldviews held within its neural network [37]. Consequently, the resulting model response diverges from a genuine perspective as it lacks the lived experience and communicative agency necessary to author and represent a worldview. Furthermore, although intended to promote ethical use, a strict adherence to behavioral guidelines presents an additional distorting influence [2] that undermines the authenticity of an already inorganic response. Especially in the political realm, where government-citizenry interactions form the basis of policy and regulation, a direct connection with the real world is a fundamental requirement for facilitating political discourse - one that GAI models inherently lack [36] [38]. Thus, any bias stemming from generative AI models poses a direct threat to the legitimacy and integrity of political deliberation, where political perceptions revolve around artificial viewpoints rather than lived experiences. The irreplaceability of humans in political discourse [38] thereby underscores the alarming nature of AI biases as they continue to permeate the social fabric.

### *D. DIFFUSED ACCOUNTABILITY*

Additionally, the inability to pinpoint responsibility for AI-generated misinformation and biases remains an unresolved issue [39]. The obfuscated practice of enigmatically deconstructing and reconstructing data [37] lends to the inability of GAI models to possess communicative agency. This absence of autonomy thereby renders any structural accountability impossible [36]. Due to the nature of LLM neural networks in forming unique and untraceable combinations [37] that potentially encompass millions of nodes, it is practically impossible to map a model response back to a specific individual responsible for a certain instance of training data [40]. The moral liability, then, is ultimately diffused across end-users who intentionally or unintentionally disseminate the AI-generated/influenced rhetoric through various avenues such as social media and news networks [36]. With responsibility spread across a wider population, each individual inherently feels a reduced sense of agency to uphold information ethics [41]. Without a definitive moral incentive to scrutinize GAI output, unsuspecting individuals will continue to exacerbate the unchecked dissemination of information, leading to a breakdown in accountability.

### *E. LIMITATIONS*

The most notable limitation of this framework is its inability to evaluate AI hallucinations holistically. Without a readily accessible heuristic algorithm that can be applied to detect AI hallucinations, identifying and pinpointing informational inaccuracies proved to be unfeasible.

Another limitation is the lack of a response specificity metric. Its incorporation requires an advanced semantic analysis that transcends lexical approaches. Such a metric may produce valuable insights into the behavior of GAI.

As with all bias evaluation frameworks, subjectivity is inevitable to some extent [52]. Primarily, the selection of weights introduces a considerable amount of human subjectivity into this methodology. Although this study strives to mitigate human partiality, zero-shot classification models are not entirely invulnerable to human biases.

## VI. CONCLUSION

In light of the consistent liberal-authoritarian biases discovered in all six evaluated LLMs, it remains clear that GAI continues to influence and skew public discourse. Its exact effect, however, is dependent on pre-existing socio-political structures within a respective geographical region; in unilateral countries such as China, political biases fuel conformity, while in bilateral environments, it becomes a polarizing influence. Given these intrinsic biases along with the diffusion of GAI responsibility and adherence to developer guidelines (and in some cases, national agendas), the normalization of AI illiteracy subjects an unsuspecting population to an elitist force overseeing the development of LLMs–whether that be AI developers or political figures. The resulting breakdown of individualistic worldviews and traditional avenues of political socialization calls for the reinforcement of AI literacy throughout society, where individuals play the role of skeptics in evaluating model responses. As GAI advances into the future, future researchers should account for response specificity and AI hallucinations to holistically determine the full extent of GAI political partisanship and subsequently propose novel measures to combat it.

## VII. REFERENCES

[1] Yong Jin Park (2021). Structural Logic of AI Surveillance and Its Normalisation in the Public Sphere, *Javnost - The Public*, 28:4, 341-357, DOI: 10.1080/13183222.2021.1955323. Accessible: https://doi.org/10.1080/13183222.2021.1955323

[2] Matthew G. Hanna, Liron Pantanowitz, Brian Jackson, Octavia Palmer, Shyam Visweswaran, Joshua Pantanowitz, Mustafa Deebajah, Hooman H. Rashidi, Ethical and Bias Considerations in Artificial Intelligence/Machine Learning, *Modern Pathology*, Volume 38, Issue 3, 2025, 100686, ISSN 0893-3952, https://doi.org/10.1016/j.modpat.2024.100686.

[3] Reznikov, Roman, PRACTICAL RECOMMENDATION OF USING GENERATIVE AI IN BUSINESS (May 01, 2024). Accessible at *SSRN*: https://ssrn.com/abstract=4851637 or http://dx.doi.org/10.2139/ssrn.4851637

[4] Bick, A., Blandin, A., Deming, D.J., 2025; The Rapid Adoption of Generative AI, *Federal Reserve Bank of St. Louis Working Paper* 2024-027. URL https://doi.org/10.20955/wp.2024.027

[5] Liao, Ruohuang. (2024). The impact of AI-generated content dissemination on social media on public sentiment. *Applied and Computational Engineering*. 90. 82-88. 10.54254/2755-2721/90/20241698. Accessible: https://www.researchgate.net/publication/385334413_The_impact_of_AI-generated_content_dissemination_on_social_media_on_public_sentiment

[6] Germani F, Spitale G, Biller-Andorno N. The Dual Nature of AI in Information Dissemination: Ethical Considerations. *JMIR AI*. 2024 Oct 15;3:e53505. Doi: 10.2196/53505. PMID: 39405099; PMCID: PMC11522648. https://pmc.ncbi.nlm.nih.gov/articles/PMC11522648/

[7] Mitra Madanchian, Hamed Taherdoost, The impact of artificial intelligence on research efficiency, *Results in Engineering*, Volume 26, 2025, 104743, ISSN 2590-1230, https://doi.org/10.1016/j.rineng.2025.104743.

[8] Holmström, J., & Carroll, N. (2025). How organizations can innovate with generative AI. *Business Horizons*, 68(5), 559–573. https://doi.org/10.1016/j.bushor.2024.02.010

[9] Tai MC. The impact of artificial intelligence on human society and bioethics. *Tzu Chi Med J*. 2020 Aug 14;32(4):339-343. doi: 10.4103/tcmj.tcmj_71_20. PMID: 33163378; PMCID: PMC7605294. Accessible: https://pmc.ncbi.nlm.nih.gov/articles/PMC7605294/

[10] Walter Y. The Future of Artificial Intelligence Will Be "Next to Normal"—A Perspective on Future Directions and the Psychology of AI Safety Concerns. *Nature Anthropology* 2024, 2, 10001. https://doi.org/10.35534/natanthropol.2024.10001

[11] Zhai, C., Wibowo, S. & Li, L.D. The effects of over-reliance on AI dialogue systems on students' cognitive abilities: a systematic review. *Smart Learn. Environ*. 11, 28 (2024). https://doi.org/10.1186/s40561-024-00316-7

[12] Artur Klingbeil, Cassandra Grützner, Philipp Schreck, Trust and reliance on AI — An experimental study on the extent and costs of overreliance on AI, *Computers in Human Behavior*, Volume 160, 2024, 108352, ISSN 0747-5632, https://doi.org/10.1016/j.chb.2024.108352.

[13] Bai, H., Voelkel, J.G., Muldowney, S. et al. LLM-generated messages can persuade humans on policy issues. *Nature Communications* 16, 6037 (2025). https://doi.org/10.1038/s41467-025-61345-5

[14] Josh A Goldstein, Jason Chao, Shelby Grossman, Alex Stamos, Michael Tomz, How persuasive is AI-generated propaganda?, *PNAS Nexus*, Volume 3, Issue 2, February 2024, pgae034, https://doi.org/10.1093/pnasnexus/pgae034

[15] Elise Karinshak, Sunny Xun Liu, Joon Sung Park, and Jeffrey T. Hancock. 2023. Working With AI to Persuade: Examining a Large Language Model's Ability to Generate Pro-Vaccination Messages. *Proc. ACM Hum.-Comput. Interact.* 7, CSCW1, Article 116 (April 2023), 29 pages. https://doi.org/10.1145/3579592

[16] Bang, Yejin & Chen, Delong & Lee, Nayeon & Fung, Pascale. (2024). Measuring Political Bias in Large Language Models: What Is Said and How It Is Said. In *Proceedings of the 62nd Annual Meeting of the Association for Computational Linguistics (Volume 1: Long Papers)*, 11142-11159. 10.18653/v1/2024.acl-long.600. Accessible: https://aclanthology.org/2024.acl-long.600/

[17] Zhu, S., Wang, W., & Liu, Y. (2024). Quite good, but not enough: Nationality bias in large language models—A case study of ChatGPT. *arXiv*. https://doi.org/10.48550/arXiv.2405.06996

[18] Huang, Po-Sen & Zhang, Huan & Jiang, Ray & Stanforth, Robert & Welbl, Johannes & Rae, Jack & Maini, Vishal & Yogatama, Dani & Kohli, Pushmeet. (2019). Reducing Sentiment Bias in Language Models via Counterfactual Evaluation. In *Findings of the Association for Computational Linguistics: EMNLP 2020*, 65-83. 10.48550/arXiv.1911.03064. Accessible: https://aclanthology.org/2020.findings-emnlp.7.pdf

[19] Marinucci, L., Mazzuca, C., & Gangemi, A. Exposing implicit biases and stereotypes in human and artificial intelligence: state of the art and challenges with a focus on gender. *AI & Soc* 38, 747–761 (2023). https://doi.org/10.1007/s00146-022-01474-3

[20] Westwood, S., Grimmer, J., & Hall, H. (2025). Measuring perceived slant in large language models through user evaluations. *Stanford Graduate School of Business*. Access: ://www.gsb.stanford.edu/gsb-box/route-download/627146

[21] Rettenberger, L., Reischl, M., & Schutera, M. (2024). Assessing political bias in large language models. *arXiv*. https://doi.org/10.48550/arXiv.2405.13041

[22] Fabio Y.S. Motoki, Valdemar Pinho Neto, Victor Rangel, Assessing political bias and value misalignment in generative artificial intelligence, *Journal of Economic Behavior & Organization*, Volume 234, 2025, 106904, ISSN 0167-2681, https://doi.org/10.1016/j.jebo.2025.106904.

[23] Patil, A., & Jadon, A. (2025). Advancing reasoning in large language models: Promising methods and approaches. *arXiv*. https://doi.org/10.48550/arXiv.2502.03671

[24] Jessica López Espejel, El Hassane Ettifouri, Mahaman Sanoussi Yahaya Alassan, El Mehdi Chouham, Walid Dahhane, GPT-3.5, GPT-4, or BARD? Evaluating LLMs' reasoning ability in zero-shot setting and performance boosting through prompts, *Natural Language Processing Journal*, Volume 5, 2023, 100032, ISSN 2949-7191, https://doi.org/10.1016/j.nlp.2023.100032.

[25] Valdemar Danry, Pat Pataranutaporn, Matthew Groh, and Ziv Epstein. 2025. Deceptive Explanations by Large Language Models Lead People to Change their Beliefs About Misinformation More Often than Honest Explanations. In Proceedings of the 2025 CHI Conference on Human Factors in Computing Systems (CHI '25). *Association for Computing Machinery*, New York, NY, USA, Article 933, 1–31. https://doi.org/10.1145/3706598.3713408

[26] Du, Y. (2025). Confirmation bias in generative AI chatbots: Mechanisms, risks, mitigation strategies, and future research directions. *arXiv*. https://doi.org/10.48550/arXiv.2504.09343

[27] Uwe Messer, How do people react to political bias in generative artificial intelligence (AI)?, *Computers in Human Behavior: Artificial Humans*, Volume 3, 2025, 100108, ISSN 2949-8821, https://doi.org/10.1016/j.chbah.2024.100108.

[28] Rozado, D. (2024). The political preferences of LLMs. *PLOS ONE* 19(7): e0306621. https://doi.org/10.1371/journal.pone.0306621

[29] Batzner, J., Stocker, V., Schmid, S., & Kasneci, G. (2024). GermanPartiesQA: Benchmarking commercial large language models for political bias and sycophancy. *arXiv*. https://doi.org/10.48550/arXiv.2407.18008

[30] Hartmann, J., Schwenzow, J., & Witte, M. (2023). The political ideology of conversational AI: Converging evidence on ChatGPT's pro-environmental, left-libertarian orientation. *arXiv*. https://doi.org/10.48550/arXiv.2301.01768

[31] Motoki, F., Pinho Neto, V. & Rodrigues, V. More human than human: measuring ChatGPT political bias. *Public Choice* 198, 3–23 (2023). https://doi.org/10.1007/s11127-023-01097-2

[32] Tavishi Choudhary. Political Bias in AI-Language Models: A Comparative Analysis of ChatGPT-4, Perplexity, Google Gemini, and Claude. *TechRxiv*. July 15, 2024. DOI: 10.36227/techrxiv.172107441.12283354/v1. Accessible: https://www.techrxiv.org/users/799951/articles/1181157-political-bias-in-ai-language-models-a-comparative-analysis-of-chatgpt-4-perplexity-google-gemini-and-claude

[33] Rozado, D. (2025). Measuring political preferences in AI systems: An integrative approach. *arXiv*. https://doi.org/10.48550/arXiv.2503.10649

[34] Elbouanani, A., Dufraisse, E., & Popescu, A. (2025). Analyzing political bias in LLMs via target-oriented sentiment classification. *arXiv*. https://doi.org/10.48550/arXiv.2505.19776

[35] Omer Yair, Gregory A. Huber, How Robust Is Evidence of Partisan Perceptual Bias in Survey Responses? A New Approach for Studying Expressive Responding, *Public Opinion Quarterly*, Volume 84, Issue 2, Summer 2020, Pages 469–492, https://doi.org/10.1093/poq/nfaa024

[36] Monti, Paolo. (2024). AI Enters Public Discourse: A Habermasian Assessment of the Moral Status of Large Language Models. *Etica e Politica*. XXVI. 61-80. Accessible: https://philpapers.org/archive/MONAEP-2.pdf

[37] Bathaee, Y. (2018). The Artificial Intelligence Black Box and the Failure of Intent and Causation. *Harvard Journal of Law and Technology.* Volume 31, No. 2. Accessible: https://jolt.law.harvard.edu/assets/articlePDFs/v31/The-Artificial-Intelligence-Black-Box-and-the-Failure-of-Intent-and-Causation-Yavar-Bathaee.pdf

[38] Kreps, Sarah & Kriner, Doug. (2023). How AI Threatens Democracy. *Journal of Democracy*. 34. 122-131. 10.1353/jod.2023.a907693. Accessible: https://www.journalofdemocracy.org/articles/how-ai-threatens-democracy/

[39] Bleher H., Braun M. Diffused responsibility: attributions of responsibility in the use of AI-driven clinical decision support systems. *AI Ethics*. 2022;2(4):747-761. doi: 10.1007/s43681-022-00135-x. Epub 2022 Jan 24. PMID: 35098247; PMCID: PMC8785388. Accessible: https://pubmed.ncbi.nlm.nih.gov/35098247/

[40] Bilal, A., Ebert, D., & Lin, B. (2025). LLMs for explainable AI: A comprehensive survey. *arXiv*. https://doi.org/10.48550/arXiv.2504.00125

[41] Beyer F., Sidarus N., Bonicalzi S., Haggard P. Beyond self-serving bias: diffusion of responsibility reduces sense of agency and outcome monitoring. *Soc Cogn Affect Neurosci*. 2017 Jan 1;12(1):138-145. doi: 10.1093/scan/nsw160. PMID: 27803288; PMCID: PMC5390744. Accessible: https://pubmed.ncbi.nlm.nih.gov/27803288/

[42] Cain, Matthew. (2015). "Political Intolerance in the 21st Century: The Role of Ideology and Emotion in Determining Intolerant Judgments". *Masters Theses*. 2339. https://thekeep.eiu.edu/theses/2339

[43] Fulay, S., Brannon, W., Mohanty, S., Overney, C., Poole-Dayan, E., Roy, D., & Kabbara, J. (2024). On the relationship between truth and political bias in language models. In Y. Al-Onaizan, M. Bansal, & Y.-N. Chen (Eds.), *Proceedings of the 2024 Conference on Empirical Methods in Natural Language Processing* (pp. 9004–9018). Association for Computational Linguistics. https://doi.org/10.18653/v1/2024.emnlp-main.508

[44] Iyengar, S., Lelkes, Y., Levendusky, M., Malhotra, N., & Westwood, S. J. (2019). The origins and consequences of affective polarization in the United States. *Annual Review of Political Science*, *22*129–146. https://doi.org/10.1146/annurev-polisci-051117-073034

[45] Geoffrey Layman, Thomas Carsey, and Juliana Menasce Horowitz, "Party Polarization in American Politics: Characteristics, Causes, and Consequences," *Annual Review of Political Science* 9 (2006): 83–110. Accessible: https://doi.org/10.1146/annurev.polisci.9.070204.105138

[46] Cyberspace Administration of China. (2023, July 13). 生成式人工智能服务管理暂行办法. 生成式人工智能服务管理暂行办法_中央网络安全和信息化委员会办公室. http://www.cac.gov.cn/2023-07/13/c_1690898327029107.htm

[47] Hui VT. Cultural Diversity and Coercive Cultural Homogenization in Chinese History. In: Phillips A, Reus-Smit C, eds. Culture and Order in World Politics. *LSE International Studies*. Cambridge University Press; 2020:93-112.

[48] Huang, L. C., & Harris, M. B. (1973). Conformity in Chinese and Americans: A Field Experiment. *Journal of Cross-Cultural Psychology*, 4(4), 427-434. https://doi.org/10.1177/002202217300400404 (Original work published 1973)

[49] Chen, Yuxin. (2025). The Accuracy and Biases of AI-Based Internet Censorship in China. *Journal of Research in Social Science and Humanities*. 4. 27-36. 10.56397/JRSSH.2025.02.05. Accessible: https://www.pioneerpublisher.com/jrssh/article/view/1223

[50] Davidson, H. (2023, February 23). "Political propaganda": China clamps down on access to ChatGPT. *The Guardian*. https://www.theguardian.com/technology/2023/feb/23/china-chatgpt-clamp-down-propaganda

[51] Ottinger, L., & Schneider, J. (2025, June 5). How to use banned US models in China. How to Use Banned US Models in China. *China Talk*. https://www.chinatalk.media/p/the-grey-market-for-american-llms

[52] Berrayana, L., Rooney, S., Garcés-Erice, L., & Giurgiu, I. (2025). Are bias evaluation methods biased? In O. Arviv, M. Clinciu, K. Dhole, R. Dror, S. Gehrmann, E. Habba, I. Itzhak, S. Mille, Y. Perlitz, E. Santus, J. Sedoc, M. Shmueli Scheuer, G. Stanovsky, & O. Tafjord (Eds.), *Proceedings of the Fourth Workshop on Generation, Evaluation and Metrics (GEM²)* (pp. 249–261). Association for Computational Linguistics. https://aclanthology.org/2025.gem-1.22/

**Dataset Attribution**

@dataset{deepseek_geopolitical_bias_dataset,
title={DeepSeek Geopolitical Bias Dataset},
author={Nitin Aravind Birur, Divyanshu Kumar, Tanay Baswa, Prashanth Harshangi, Sahil Agarwal},
year={2025},
url={https://huggingface.co/datasets/enkryptai/deepseek-geopolitical-bias-dataset},
description={A dataset for analyzing bias and censorship in LLM responses to geopolitical questions.}
}

**NATHAN JUNZI CHEN** was born in West Covina, CA, USA, in 2008. In the summer of 2023, he attended Collin College in Allen, TX, USA, where he received a specialized education in cyber psychology. He currently pursues the cybersecurity pathway at Troy High School in Fullerton, CA, USA, with a primary focus on Windows OS Security. He is interested in exploring the intersection between the human mind and artificial intelligence and aspires to partake in the development of human-aligned AI. He served as a paralegal intern in the summer of 2025, during which time he examined case law and investigated the legal gray zones surrounding generative AI technologies.